# An Evaluation of Naive Bayesian Anti-Spam Filtering

Ion Androutsopoulos, John Koutsias, Konstantinos V. Chandrinos, George Paliouras
and Constantine D. Spyropoulos

Software and Knowledge Engineering Laboratory
National Centre for Scientific Research "Demokritos"
153 10 Ag. Paraskevi, Athens, Greece
phone: +30-1-6503197
fax: +30-1-6532175
E-mail: {ionandr, jkoutsi, kostel, paliourg, costass}@iit.demokritos.gr

**Abstract**

It has recently been argued that a Naive Bayesian classifier can be used to filter unsolicited bulk e-mail ("spam"). We conduct a thorough evaluation of this proposal on a corpus that we make publicly available, contributing towards standard benchmarks. At the same time we investigate the effect of attribute-set size, training-corpus size, lemmatization, and stop-lists on the filter's performance, issues that had not been previously explored. After introducing appropriate cost-sensitive evaluation measures, we reach the conclusion that additional safety nets are needed for the Naive Bayesian anti-spam filter to be viable in practice.

## 1 Introduction

Unsolicited bulk e-mail, electronic messages posted blindly to thousands of recipients, is becoming alarmingly common. Although most users find these postings – called "spam" – annoying and delete them immediately, the low cost of e-mail is a strong incitement for direct marketers advertising anything from vacations to get-rich schemes. A 1997 study (Cranor & LaMacchia, 1998) found that 10% of the incoming e-mail to a corporate network was spam. Apart from wasting time, spam costs money to users with dial-up connections, wastes bandwidth, and may expose under-aged recipients to unsuitable (e.g. pornographic) content.

Some anti-spam filters are already available.[1] These rely mostly on manually constructed pattern-matching rules that need to be tuned to each user's incoming messages, a task requiring time and expertise. Furthermore, the characteristics of spam (e.g. products advertised, frequent terms) change over time, requiring the rules to be maintained. A system that would learn *automatically* to separate spam from other "legitimate" messages would, therefore, present significant advantages.

Several machine learning algorithms have been applied to text categorization (e.g. Apte & Damerau, 1994; Lewis, 1996; Dagan et al., 1997; see Sebastiani, 1999, for a survey). These algorithms learn to classify documents into fixed categories, based on their content, after being trained on manually categorized documents. Algorithms of this kind have also been used to thread e-mail (Lewis & Knowles, 1997), classify e-mail into folders (Cohen, 1996; Payne & Edwards, 1997), identify interesting news articles (Lang, 1995), etc. To the best of our knowledge, however, only one attempt has ever been made to apply a machine learning algorithm to anti-spam filtering (Sahami et al., 1998).

Sahami et al. trained a Naive Bayesian classifier (Duda & Hart, 1973; Mitchell 1997) on manually categorized legitimate and spam messages, reporting impressive precision and recall on unseen messages. It may be surprising that text categorization can be effective in anti-spam filtering: unlike other text categorization tasks, it is the *act* of blindly mass-mailing a message that makes it spam, not its actual content. Nevertheless, it seems that the language of spam constitutes a distinctive genre, and that spam messages are often about topics rarely mentioned in legitimate messages, making it possible to train a text classifier for anti-spam filtering.

---

[1] See, for example, http://www.tucows.com. Consult http://www.cauce.org, http://www.junkemail.org, and http://spam.abuse.net for related resources and legal issues.

Text categorization research has benefited from publicly available manually categorized document collections, like the Reuters corpus (Lewis, 1992), that have been used as benchmarks. Creating similar resources for anti-spam filtering is not straightforward, because a user's incoming e-mail stream cannot be made public without violating his/her privacy. A useful approximation of such a stream, however, can be made by mixing spam messages with messages extracted from spam-free public archives of mailing lists. Towards that direction, we test Sahami et al.'s approach on a mixture of spam messages and messages sent via the Linguist list,[2] a moderated (hence, spam-free) list about the profession and science of linguistics. The resulting corpus, dubbed Ling-Spam, is made publicly available for others to use as a benchmark.[3]

The Linguist messages are, of course, more topic-specific than most users' incoming e-mail. They are less standardized, however, than one might expect (e.g. they contain job postings, software availability announcements, even flame-like responses), to the extent that useful preliminary conclusions about anti-spam filtering of a user's incoming e-mail can be reached with Ling-Spam, at least until better public corpora become available. With a more direct interpretation, our experiments can be seen as a study on anti-spam filters for open unmoderated mailing lists or newsgroups.

Unlike Sahami et al., we use ten-fold cross-validation which makes our results less prone to random variation. Our experiments also shed more light on the behavior of Naive Bayesian anti-spam filtering by investigating the effect of attribute-set size, training-corpus size, lemmatization, and stop-lists, issues not covered by Sahami et al.'s study. Furthermore, we show how evaluation measures that incorporate a decision-theoretic notion of cost can be employed. Our results confirm Sahami et al.'s high precision and recall. A cost-sensitive evaluation, however, suggests that complementary safety nets are needed for the Naive Bayesian filter to be viable.

Section 2 discusses Naive Bayesian classification; section 3 lists Sahami et al.'s results; section 4 describes our filtering system, the Ling-Spam corpus, and our results; section 5 introduces cost-sensitive evaluation measures; and section 6 concludes.

## 2  Naive Bayesian classification

Each message is represented by a vector $\vec{x} = \langle x_1, x_2, x_3, \ldots, x_n \rangle$, where $x_1, \ldots, x_n$ are the values of attributes $X_1, \ldots, X_n$. Following Sahami et al., we use binary attributes: $X_i = 1$ if some characteristic represented by $X_i$ is present in the message; otherwise $X_i = 0$. In our experiments, attributes correspond to words, i.e. each attribute shows if a particular word (e.g. "adult") is present. To select among all possible attributes, we follow Sahami et al. and compute the mutual information ($MI$) of each candidate attribute $X$ with the category-denoting variable $C$:

$$MI(X; C) = \sum_{x \in \{0,1\}, c \in \{spam, legitimate\}} P(X = x, C = c) \cdot \log \frac{P(X = x, C = c)}{P(X = x) \cdot P(C = c)}$$

The attributes with the highest $MI$s are selected. Probabilities are estimated as frequency ratios from the training corpus (see Mitchell, 1996, for better estimators that we plan to incorporate in future).

From Bayes' theorem and the theorem of total probability, given the vector $\vec{x} = \langle x_1, \ldots, x_n \rangle$ of a document $d$, the probability that $d$ belongs to category $c$ is:

$$P(C = c \mid \vec{X} = \vec{x}) = \frac{P(C = c) \cdot P(\vec{X} = \vec{x} \mid C = c)}{\sum_{k \in \{spam, legitimate\}} P(C = k) \cdot P(\vec{X} = \vec{x} \mid C = k)}$$

The probabilities $P(\vec{X} \mid C)$ are practically impossible to estimate directly (the possible values of $\vec{X}$ are too many, and there are data-sparseness problems). The Naive Bayesian classifier makes the simplifying assumption that $X_1, \ldots, X_n$ are conditionally independent given the category $C$. Then:

---

[2] Archived at http://listserv.linguistlist.org/archives/linguist.html.
[3] The Ling-Spam corpus is available from http://www.iit.demokritos.gr/~ionandr/publications.htm.

$$P(C = c \mid \vec{X} = \vec{x}) = \frac{P(C = c) \cdot \prod_{i=1}^{n} P(X_i = x_i \mid C = c)}{\sum_{k \in \{spam, legitimate\}} P(C = k) \cdot \prod_{i=1}^{n} P(X_i = x_i \mid C = k)}$$

where $P(X_i \mid C)$ and $P(C)$ can be easily estimated as relative frequencies from the training corpus. Several studies have found the Naive Bayesian classifier to be surprisingly effective (Langley et al., 1992; Domingos & Pazzani, 1996), despite the fact that its independence assumption is usually over-simplistic.

Mistakenly blocking a legitimate message (classifying it as spam) is generally more severe than letting a spam message pass the filter (classifying a spam message as legitimate). Let $L \rightarrow S$ and $S \rightarrow L$ denote the two error types. Assuming that $L \rightarrow S$ is $\lambda$ times more costly than $S \rightarrow L$, we classify a message as spam if:

$$\frac{P(C = spam \mid \vec{X} = \vec{x})}{P(C = legitimate \mid \vec{X} = \vec{x})} > \lambda$$

To the extent that the independence assumption holds and the probability estimates are accurate, a classifier adopting this criterion achieves optimal results (Duda & Hart, 1973). In our case, $P(C = spam \mid \vec{X} = \vec{x}) = 1 - P(C = legitimate \mid \vec{X} = \vec{x})$, which leads to an alternative reformulation of the criterion:

$$P(C = spam \mid \vec{X} = \vec{x}) > t, \text{ with } t = \frac{\lambda}{1 + \lambda}, \lambda = \frac{t}{1 - t}$$

Sahami et al. set the threshold $t$ to 0.999 ($\lambda = 999$); i.e. blocking a legitimate message is as bad as letting 999 spam messages pass the filter. Such a high value of $\lambda$ is reasonable when blocked messages are discarded without further processing, as most users would consider losing a legitimate message unacceptable. Alternative configurations are possible, however, where lower values of $\lambda$ are reasonable. Instead of deleting a blocked message, it could be returned to the sender, with a request to re-send it to a private un-filtered e-mail address of the recipient (see also Hall, 1998). The private address would never be advertised (e.g. on web pages), making it unlikely to receive spam directly; and the request to re-send could include a frequently changing riddle (e.g. "Include in the subject the capital of France.") to ensure that replies are not sent by spam-generating robots. In that case, $\lambda = 9$ ($t = 0.9$) seems reasonable: blocking a legitimate message is penalized mildly more than letting a spam message pass, to model the fact that re-sending a blocked message involves more work (by the sender) than manually deleting a spam message. Even $\lambda = 1$ ($t = 0.5$) may be acceptable, if the recipient does not care about extra work imposed on the sender.

## 3   Previous results

Table 1 summarizes Sahami et al.'s results. If $n_{L \rightarrow S}$ and $n_{S \rightarrow L}$ are the numbers of $L \rightarrow S$ and $S \rightarrow L$ errors, and $n_{L \rightarrow L}$, $n_{S \rightarrow S}$ are the numbers of correctly treated legitimate and spam messages, then spam recall ($SR$) and spam precision ($SP$) are:

$$SR = \frac{n_{S \rightarrow S}}{n_{S \rightarrow S} + n_{S \rightarrow L}} \qquad SP = \frac{n_{S \rightarrow S}}{n_{S \rightarrow S} + n_{L \rightarrow S}}$$

In the second experiment of table 1, candidate attributes included not only word-attributes, but also attributes showing if particular hand-picked phrases (e.g. "be over 21") were present. In the third and fourth experiments, non-textual candidate attributes were added, showing if messages had manually chosen properties (e.g. attachments). Sahami et al.'s phrasal and non-textual attributes introduce a manual configuration stage, as one has to select manually phrases and non-textual characteristics to be treated as candidate attributes. Since our target was to explore fully automatic anti-spam filtering, we have limited ourselves to word-attributes.

Table 1: Resuls of Sahami et al. (500 attributes, threshold = 0.999, $\lambda = 999$ )

| Attributes | Total Messages | Testing Messages | % Spam | Spam Precision | Spam Recall |
|---|---|---|---|---|---|
| words only | 1789 | 251 | 88.2% | 97.1% | 94.3% |
| words + phrases | 1789 | 251 | 88.2% | 97.6% | 94.3% |
| words + phrases + non-textual | 1789 | 251 | 88.2% | 100.0% | 98.3% |
| words + phrases + non-textual | 2815 | 222 | ~20% | 92.3% | 80.0% |

## 4  Experiments with Ling-Spam

Our experiments were all performed on the Ling-Spam corpus, which consists of:

- 2412 Linguist messages, obtained by randomly downloading digests from the archives, separating their messages, and removing text added by the list's server.

- 481 spam messages, received by the first author. Attachments, HTML tags, and duplicate spam messages received on the same day were not included.

Spam is 16.6% of the corpus, a figure close to the spam rates of the authors, Sahami et al.'s fourth experiment, and rates reported elsewhere (Cranor & LaMacchia, 1998).

Our implementation of the Naive Bayesian filter (developed on GATE), includes a lemmatizer that converts each word to its base form, and a stop-list that removes from messages the 100 most frequent words of the British National Corpus (BNC).[4] The two modules can be enabled or disabled, allowing their effect to be measured.

To reduce random variation, ten-fold cross-validation was employed, and averaged scores are reported. In a first series of experiments, the number of retained attributes (highest $MI$) ranged from 50 to 700 by 50, for all combinations of enabled/disabled lemmatizer and stop-list. Three thresholds were tried: $t = 0.999$ ( $\lambda = 999$ ), $t = 0.9$ ( $\lambda = 9$ ), and $t = 0.5$ ( $\lambda = 1$ ). As discussed in section 2, these represent three scenarios: deleting blocked messages; issuing a re-send request and accounting for the sender's extra work; and issuing a re-send request ignoring the sender's extra work.

Figures 1 – 3 show that the filter achieved impressive spam recall and precision at all three thresholds, verifying in that sense the findings of Sahami et al. In all cases, lemmatization seems to improve results. The stop-list has a positive effect for $\lambda = 1$ and $\lambda = 9$, but its effect looks negligible for $\lambda = 999$. Without a single evaluation measure, however, to be used instead of spam precision and recall, it is difficult to check if the effects of the lemmatizer and the stop-list are statistically significant.

For $\lambda = 999$, blocking a legitimate message is much more severe than letting a spam message pass the filter. Hence, it seems reasonable to assume that the "best" configuration is the one that maximizes spam precision. This is achieved with 300 attributes and the lemmatizer enabled (100% spam precision, 63% spam recall; here, the effect of the stop-list is negligible). For $\lambda = 1$ and $\lambda = 9$, however, it is hard to tell which configuration (combination of precision and recall) is best. Again, a single measure is needed; and it must be sensitive to our cost. We discuss this next.[5]

---

[4] GATE, including the lemmatizer, is available from http://www.dcs.shef.ac.uk/research/groups/nlp. BNC frequency lists are available from ftp://ftp.itri.bton.ac.uk/pub/bnc.
[5] The F-measure, used in information retrieval and extraction to combine recall and precision, is unsuitable to our purposes, because its weighting factor cannot be easily related to our notion of cost.

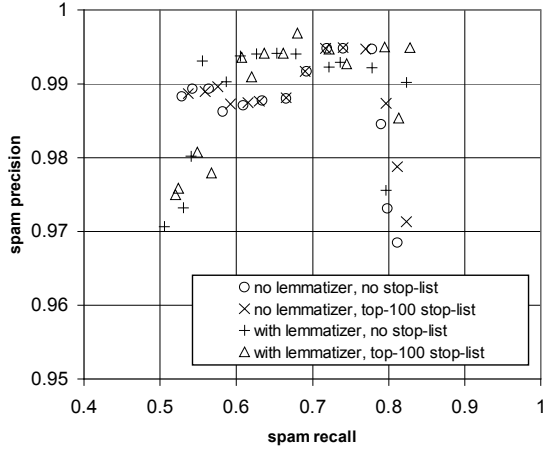# 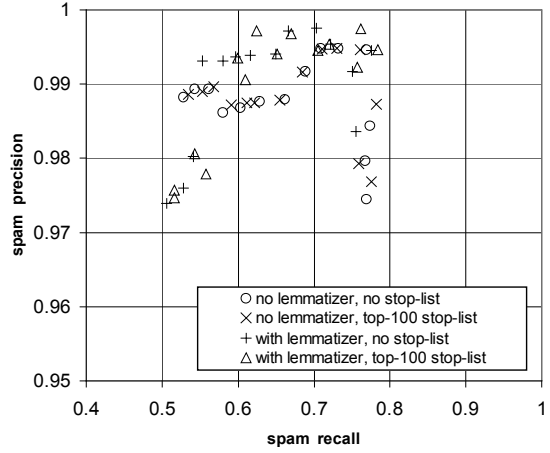

Figure 1: Spam precision and recall at $t = 0.5$ ($\lambda = 1$)

Figure 2: Spam precision and recall at $t = 0.9$ ($\lambda = 9$)

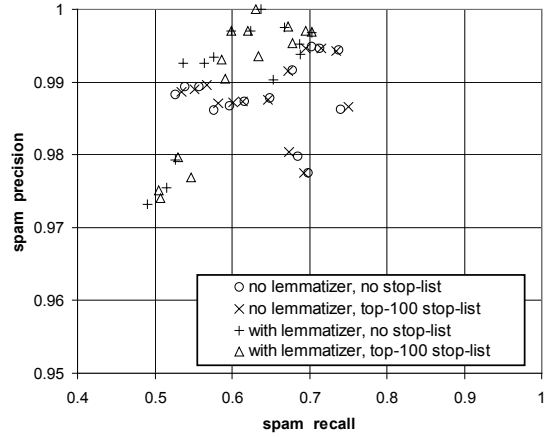

Figure 3: Spam precision and recall at $t = 0.999$ ($\lambda = 999$)

## 5 Cost-sensitive evaluation measures

In classification tasks, two commonly used evaluation measures are accuracy ($Acc$) and error rate ($Err = 1 - Acc$). In our case:

$$Acc = \frac{n_{L \to L} + n_{S \to S}}{N_L + N_S} \qquad Err = \frac{n_{L \to S} + n_{S \to L}}{N_L + N_S}$$

$N_L$ and $N_S$ are the numbers of legitimate and spam messages to be classified.

Accuracy and error rate assign equal weights to the two error types ($L \to S$ and $S \to L$). When selecting the threshold of the classifier (section 2), however, we assumed that $L \to S$ is $\lambda$ times more costly than $S \to L$. To make accuracy and error rate sensitive to this cost, we treat each legitimate message as if it were $\lambda$ messages: when a legitimate message is misclassified, this counts as $\lambda$ errors; and when it is classified correctly, this counts as $\lambda$ successes. This leads to weighted accuracy ($WAcc$) and weighted error rate ($WErr = 1 - WAcc$):

$$WAcc = \frac{\lambda \cdot n_{L \to L} + n_{S \to S}}{\lambda \cdot N_L + N_S} \qquad WErr = \frac{\lambda \cdot n_{L \to S} + n_{S \to L}}{\lambda \cdot N_L + N_S}$$

Table 2: Results on Ling-Spam for best no. of attributes (2893 total messages, 16.6% spam, 10-fold cross validation, attributes ranging from 50 to 700 by a step of 50)

| Filter Configuration | $\lambda$ | No. of attrib. | Spam Recall | Spam Precision | Weighted Accuracy | Baseline W. Acc. | TCR |
|---|---|---|---|---|---|---|---|
| (a) bare | 1 | 50 | 81.10% | 96.85% | 96.408% | 83.374% | 4.63 |
| (b) stop-list | 1 | 50 | 82.35% | 97.13% | 96.649% | 83.374% | 4.96 |
| (c) lemmatizer | 1 | 100 | 82.35% | 99.02% | 96.926% | 83.374% | 5.41 |
| (d) lemmatizer + stop-list | 1 | 100 | 82.78% | 99.49% | 97.064% | 83.374% | 5.66 |
| (a) bare | 9 | 200 | 76.94% | 99.46% | 99.419% | 97.832% | 3.73 |
| (b) stop-list | 9 | 200 | 76.11% | 99.47% | 99.401% | 97.832% | 3.62 |
| (c) lemmatizer | 9 | 100 | 77.57% | 99.45% | 99.432% | 97.832% | 3.82 |
| (d) lemmatizer + stop-list | 9 | 100 | 78.41% | 99.47% | 99.450% | 97.832% | 3.94 |
| (a) bare | 999 | 200 | 73.82% | 99.43% | 99.912% | 99.980% | 0.23 |
| (b) stop-list | 999 | 200 | 73.40% | 99.43% | 99.912% | 99.980% | 0.23 |
| (c) lemmatizer | 999 | 300 | 63.67% | 100.00% | 99.993% | 99.980% | 2.86 |
| (d) lemmatizer + stop-list | 999 | 300 | 63.05% | 100.00% | 99.993% | 99.980% | 2.86 |

When using accuracy or error rate (weighted or not), it is important to compare to a simplistic "baseline" approach, to avoid misinterpreting the often high accuracy and low error rate scores. As baseline, we use the case where no filter is present: legitimate messages are (correctly) never blocked, and spam messages (mistakenly) always pass the filter. The weighted accuracy and error rate of the baseline are:

$$WAcc^b = \frac{\lambda \cdot N_L}{\lambda \cdot N_L + N_S} \qquad WErr^b = \frac{N_S}{\lambda \cdot N_L + N_S}$$

To compare easily with the baseline, we introduce the total cost ratio ($TCR$):

$$TCR = \frac{WErr^b}{WErr} = \frac{N_S}{\lambda \cdot n_{L \to S} + n_{S \to L}}$$

Greater $TCR$ indicates better performance. For $TCR < 1$, not using the filter is better. If cost is proportional to wasted time, $TCR$ measures how much time is wasted to delete manually all spam messages when no filter is present ($N_S$), compared to the time wasted to delete manually any spam messages that passed the filter ($n_{S \to L}$) plus the time needed to recover from mistakenly blocked legitimate messages ($\lambda \cdot n_{L \to S}$).

Table 2 lists spam recall, spam precision, weighted accuracy, baseline weighted accuracy, and $TCR$, for various configurations of the filter, and for the number of attributes that led to the highest $TCR$ with each configuration. Figures 4 – 6 show $TCR$ for different numbers of attributes, and $\lambda = 1$, 9, 999. In all cases, ten-fold cross validation was used, and average $WAcc$ is reported. $TCR$ is computed as $WErr^b$ divided by the average $WErr$. Increasing the number of attributes beyond a certain point generally degrades performance, because attributes with low $MI$ do not discriminate well between the two categories.

At all three $\lambda$ values, the highest $TCR$ scores were obtained with the lemmatizer enabled. The stop-list had an additional positive effect for $\lambda = 1$ and $\lambda = 9$, but not for $\lambda = 999$. The differences, however, are not always statistically significant. For $\lambda = 1$, paired single-tailed t-tests on $WAcc$ between all filter configurations of table 2 confirm only that configurations (b) and (d) are better than (a) at $p < 0.05$. All four configurations, however, are significantly better than the baseline at $p < 0.01$. For $\lambda = 9$, none of

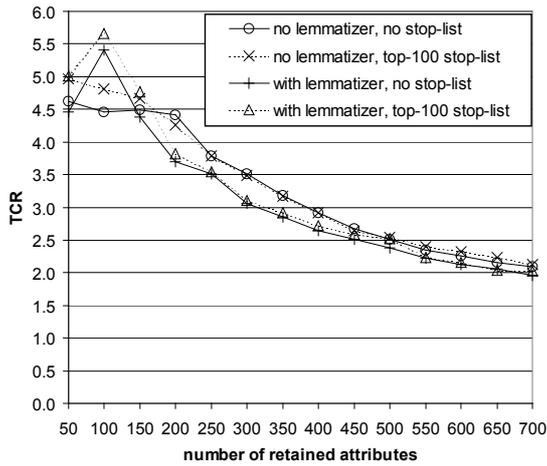
Figure 4: *TCR* at $t = 0.5$ ($\lambda = 1$)

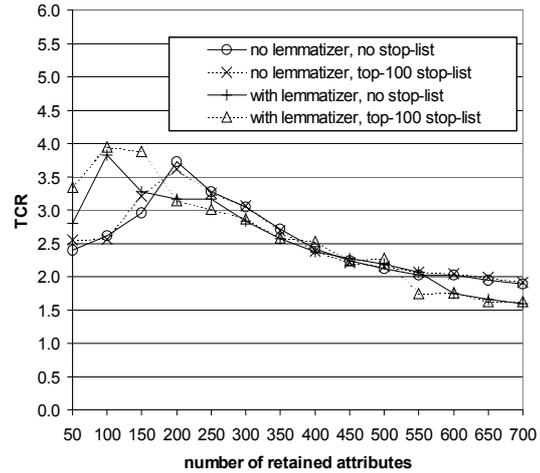
Figure 5: *TCR* at $t = 0.9$ ($\lambda = 9$)

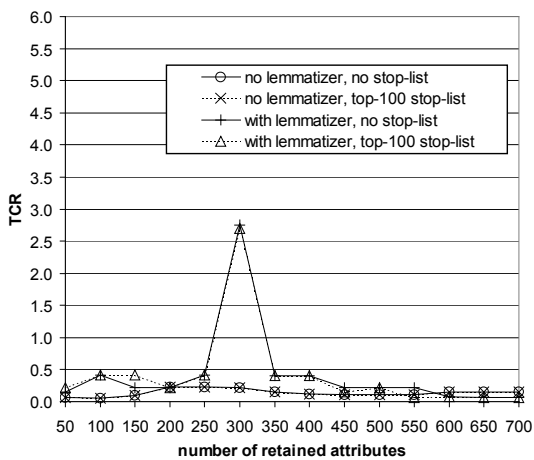
Figure 6: *TCR* at $t = 0.999$ ($\lambda = 999$)

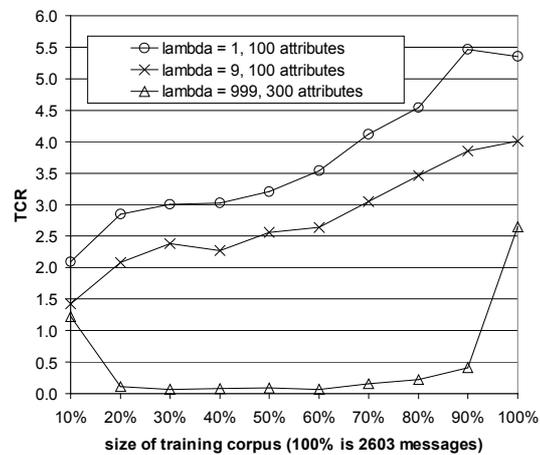
Figure 7: *TCR* for variable training corpus size, with lemmatizer and stop-list

the hypotheses of table 2, e.g. that configuration (d) is better than (a), are statistically significant at $p < 0.05$, but all configurations are, again, significantly better than the baseline at $p < 0.01$. For $\lambda = 999$, the filter achieves $TCR > 1$ only with the lemmatizer enabled. The stop-list has essentially no effect, and both configurations (c) and (d) are significantly better than the baseline at $p < 0.01$.

Overall, for $\lambda = 1$ and $\lambda = 9$ the filter demonstrates a stable behavior, with *TCR* constantly greater than 1. For $\lambda = 999$, however, the filter achieves $TCR > 1$ only for one particular number of attributes (300), because the $L \rightarrow S$ error is penalized so heavily that a single blocked legitimate message is enough for $WAcc^b$ to exceed $WAcc$ (the filter makes no such error at 300 attributes). In a real application, it is unlikely that one would be able to pin-point precisely the optimal number of attributes, which casts doubts over the applicability of the filter for $\lambda = 999$.

Even more worrying, for $\lambda = 999$, are the results of a second series of experiments we performed, this time varying the size of the training corpus. At every ten-fold repetition, Ling-Spam was divided into ten parts, with one part reserved for testing. From each one of the remaining nine parts, only $x$% was used for training, with $x$ ranging from 10 to 100 by 10. Figure 7 shows the resulting *TCR* scores for $\lambda = 1, 0.9, 0.999$. All experiments were conducted with the lemmatizer and stop-list enabled, and with the best numbers of attributes, as in table 2.

Unlike $\lambda = 1$ and $\lambda = 9$, for $\lambda = 999$ the filter reached $TCR > 1$ only with 100% of the training corpus, and one cannot easily assume that $TCR$ would remain $> 1$ given more training. (We attribute the initial peak of $TCR$ to the fact that with very little training the classifier tends to classify all messages into the most frequent category, legitimate, which protects it from making a costly $L \rightarrow S$ error). These findings suggest that when $\lambda = 999$, the filter is not safe enough to use.

## 6 Conclusions

Our cost-sensitive evaluation suggests that, despite its high spam recall and precision, the Naive Bayesian filter is not viable when blocked messages are deleted (a situation we modelled with $\lambda = 999$). With additional safety nets, however, like re-sending to private addresses, the cost of blocking a legitimate message is lower (we used $\lambda = 1$ and $\lambda = 9$), and the filter has a stable significant positive contribution.

We plan to implement anti-spam filters based on alternative machine learning algorithms, and compare them to the Naive Bayesian filter. We expect automatic anti-spam filtering to become an important member of an emerging family of junk-filtering tools for the Internet, which will include tools to remove advertisements (Kushmerick, 1999), and block hostile or pornographic material (Forsyth, 1996; Spertus, 1997).